\documentclass{SCIS2023}

\usepackage{threeparttable}
\usepackage{tabularx}
\usepackage{booktabs}
\usepackage{verbatim}
\usepackage{amsopn}
\usepackage{amsmath}
\usepackage{multirow}
\usepackage{colortbl}
\usepackage{graphicx}
\usepackage{wrapfig}

\usepackage{xspace}

\makeatletter
\DeclareRobustCommand\onedot{\futurelet\@let@token\@onedot}
\def\@onedot{\ifx\@let@token.\else.\null\fi\xspace}

\makeatother
\usepackage{epstopdf}

\usepackage[dvipsnames]{xcolor}

\usepackage{colortbl}

\usepackage[utf8]{inputenc} 
\usepackage[T1]{fontenc}    
\usepackage{url}            
\usepackage{booktabs}       
\usepackage{amsfonts}       
\usepackage{nicefrac}       
\usepackage{microtype}      

\usepackage{pifont}
\usepackage{subfig}
\usepackage{multirow}
\usepackage{graphicx}
\usepackage{wrapfig}
\usepackage{booktabs}
\usepackage{anyfontsize}
\usepackage{makecell}
\usepackage{amsthm,amsmath,amssymb}
\usepackage{mathrsfs}

\usepackage{bm}
\usepackage{bbm}
\usepackage{tikz}

\usepackage{xcolor}
\usepackage{xspace}
\definecolor{citecolor}{RGB}{75, 166, 154}
\hypersetup{
    colorlinks=true,
    linkcolor=red,
    filecolor=black,
    citecolor=blue,
    urlcolor=red,
}

\makeatletter
\DeclareRobustCommand\onedot{\futurelet\@let@token\@onedot}
\def\@onedot{\ifx\@let@token.\else.\null\fi\xspace}

\def\eg{\emph{e.g}\onedot} 
\def\ie{\emph{i.e}\onedot} 
 
\def\etc{\emph{etc}\onedot}

\makeatother

\definecolor{Gray}{gray}{0.94}
\definecolor{liGray}{gray}{0.5}
\definecolor{LightCyan}{rgb}{0.88,1,1}
\newcommand{\method}{\texttt{UniAnimate}\xspace}

\newcommand{\tocite}[1]{\textcolor{red}{[TO CITE]}}

\newlength\savewidth\newcommand\shline{\noalign{\global\savewidth\arrayrulewidth
  \global\arrayrulewidth 1pt}\hline\noalign{\global\arrayrulewidth\savewidth}}

\begin{document}
\ArticleType{RESEARCH PAPER}
\Year{}
\Month{}
\Vol{}
\No{}
\DOI{}
\ArtNo{}
\ReceiveDate{}
\ReviseDate{}
\AcceptDate{}
\OnlineDate{}

\title{UniAnimate: Taming Unified Video Diffusion Models for Consistent Human Image Animation}

%

\author[1]{Xiang WANG}{}
\author[2]{Shiwei ZHANG}{}
\author[1]{Changxin GAO}{}
\author[2]{Jiayu WANG}{}

\author[3]{Xiaoqiang ZHOU}{}

\author[2]{\\ Yingya ZHANG}{}
\author[1]{Luxin YAN}{}
\author[1]{Nong SANG}{nsang@hust.edu.cn}


\AuthorMark{Xiang WANG}

\AuthorCitation{Xiang WANG, Shiwei ZHANG, Changxin GAO, et al}


\address[1]{Key Laboratory of Ministry of Education for Image Processing and Intelligent Control,\\ School of Artificial
Intelligence and Automation, \\ 
Huazhong University of Science and Technology, Wuhan {\rm 430074}, China}
\address[2]{Alibaba Group, Hangzhou {\rm 310052}, China}
\address[3]{University of Science and Technology of China, Hefei {\rm 230026}, China}


\abstract{

Recent diffusion-based human image animation techniques have demonstrated impressive success in synthesizing videos that faithfully follow a given reference identity and a sequence of desired movement poses. 
Despite this, there are still two limitations: \emph{i)} an extra reference model is required to align the identity image with the main video branch, which significantly increases the optimization burden and model parameters; \emph{ii)} the generated video is usually short in time (\eg, 24 frames), hampering practical applications.
To address these shortcomings, we present a \method framework to enable efficient and long-term human video generation. 
First, to reduce the optimization difficulty and ensure temporal coherence, we map the reference image  along with the posture guidance and noise video into a common feature space by incorporating a unified video diffusion model.
Second, we propose a unified noise input that supports random noised input as well as first frame conditioned input, which enhances the ability to generate long-term video.
Finally, to further efficiently handle long sequences, we explore an alternative temporal modeling architecture based on state space model to replace the original computation-consuming temporal Transformer.
Extensive experimental results indicate that \method achieves superior synthesis results over existing state-of-the-art counterparts in both quantitative and qualitative evaluations.
Notably, \method can even generate highly consistent one-minute videos by iteratively employing the first frame conditioning strategy.
Code and models will be publicly available. Project page: \url{https://unianimate.github.io/}.
}


\maketitle
\section{Introduction}

Human image animation~\cite{yang2018pose,zablotskaia2019dwnet} is an attractive and challenging task that aims to generate lifelike and high-quality videos in accordance with the input reference image and target pose sequence.
This task has made unprecedented progress and showcased the potential for broad applications~\cite{Animateanyone,magicanimate,magicdance,jiang2022text2human} with the rapid advancement of video generation methods~\cite{cogvideo,magicanimate,make-a-video,imagenvideo,luo2023videofusion,zhang2023i2vgen,mocogan,ma2023dreamtalk}, especially the iterative evolution of generative models~\cite{DDPM,guo2023animatediff,videocomposer,chen2023videocrafter1,tft2v,wang2023videolcm,zhang2023controlvideo,zhao2023controlvideo,wang2020g3an}.

\begin{figure}[!t]
    \centering
    \includegraphics[width=0.97\linewidth]{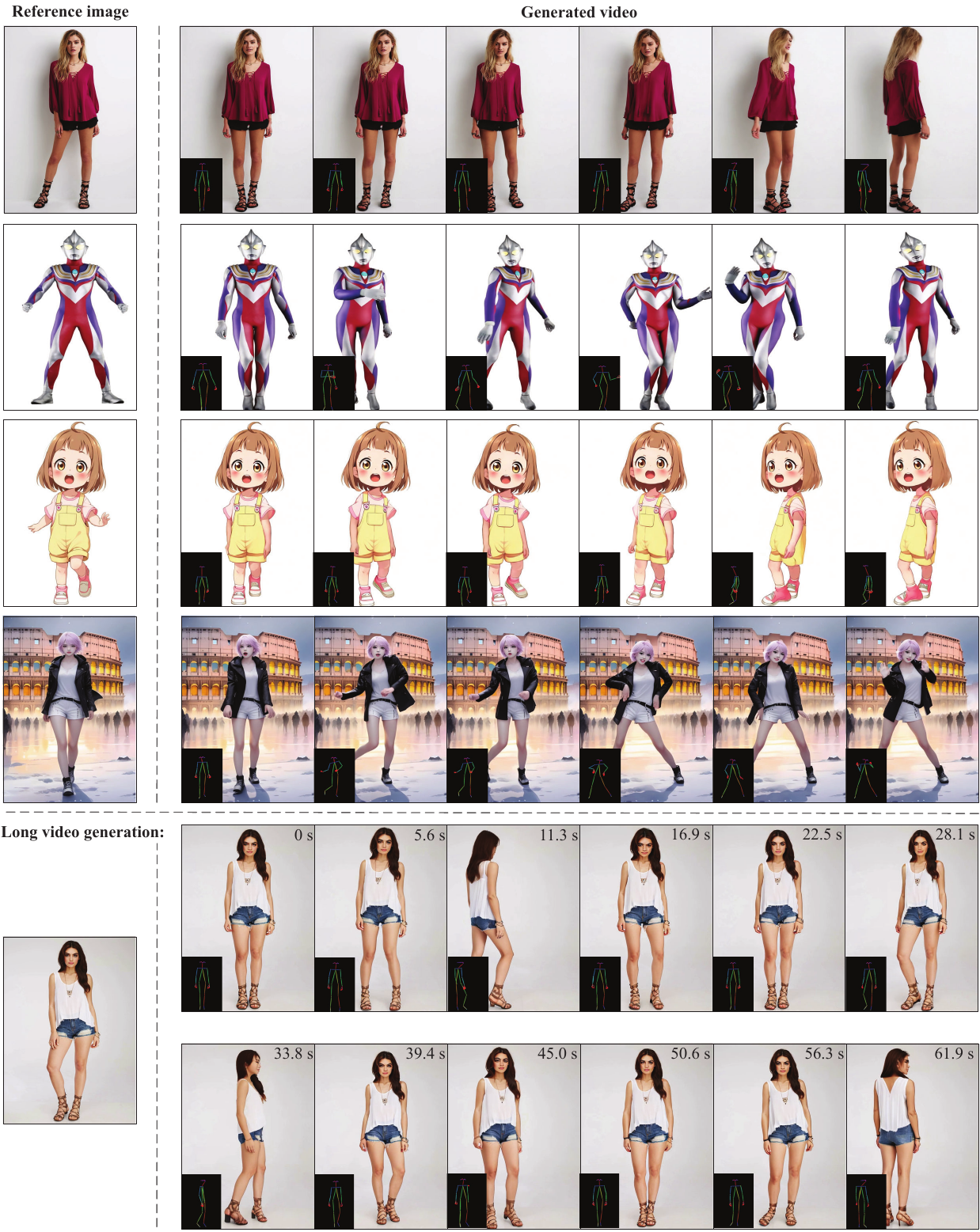}
    \vspace{-3mm}
    \caption{Example videos synthesized by the proposed \method. 
    Given a reference image and a target pose sequence, \method can generate temporally consistent and high-quality character videos that seamlessly adhere to the input conditional guidance. Note that our method is not trained on any cartoon character dancing videos, displaying excellent cross-domain transfer capability.
    In addition, by iteratively employing the first frame conditioning strategy, \method can generate high-fidelity one-minute videos. 
        }
    \label{fig:example_video}
    \vspace{-3mm}
\end{figure}

Existing methods can be broadly categorized into two groups. The first group~\cite{zablotskaia2019dwnet,yang2018pose,yu2023bidirectionally,zhang2022exploring} usually  leverages  intermediate pose-guided representation to warp the reference appearance and subsequently
utilizes a Generative Adversarial Network (GAN)~\cite{goodfellow2014generative} for plausible frame prediction conditioning on previously warped subjects.
However, GAN-based approaches generally suffer from training instability and poor generalization issues~\cite{disco,Animateanyone}, resulting in non-negligible artifacts and inter-frame jitters.
The second group~\cite{disco,Animateanyone,magicdance,magicanimate,champ,ma2024follow,karras2023dreampose} employs diffusion models to synthesize photo-realistic videos.
For instance, Disco~\cite{disco} disentangles the control signals into three conditions, \ie, subjects, backgrounds, and dance moves, and applies a ControlNet-like architecture~\cite{controlnet} for holistic background modeling and human pose transfer.
Animate Anyone~\cite{Animateanyone} and MagicAnimate~\cite{magicanimate} utilize a 3D-UNet model~\cite{VideoLDM} for video denoising and exploit an additional reference network mirrors from the main 3D-UNet branch, excluding temporal Transformer modules, to extract reference image features for appearance alignment. To encode target pose information, a lightweight pose encoder is also utilized to capture desired motion characterizations.
These methods inherit the advantages of stable training and strong transferable capabilities of diffusion models, demonstrating superior performance to GAN-based approaches~\cite{disco,Animateanyone,magicanimate}.

Despite these advancements, the existing diffusion-based methods still have two limitations: \emph{i)} they require an extra reference network to encode reference image features and align them with the main branch of the 3D-UNet, resulting in increased training difficulty and model parameter count; \emph{ii)} they usually employ temporal Transformers to model the temporal information, but Transformers require quadratic computations in the temporal dimension, which limits the generated video length. Typical methods~\cite{Animateanyone,champ} can only generate 24 frames, restricting practical deployment. Although the slide window strategy~\cite{magicanimate} that employs temporally overlapped local windows to synthesize videos and average the intersection parts is able to generate longer videos, we empirically observed that there are usually non-smooth transitions and appearance inconsistencies at the segment connections with the reference image. 

To address the aforementioned limitations, we propose the \method framework for consistent human image animation. 
Specifically,
we leverage a unified video diffusion model to simultaneously handle the reference image and noised video, facilitating feature alignment and ensuring temporally coherent video generation. Additionally, to generate smooth and continuous video sequences, we design a unified noised input that allows random noised video or first frame conditioned video as input for video synthesis.
The first frame conditioning strategy can generate subsequent frames based on the final frame of the previously generated video, ensuring smooth transitions. 
Moreover, to alleviate the constraints of generating long videos at once, we utilize temporal Mamba~\cite{mamba,Visionmamba,li2024videomamba} to replace the original temporal Transformer, significantly improving the efficiency.
By this means, \method can enable highly consistent human image animation and is able to synthesize long-term videos with smooth transitions, as displayed in Figure~\ref{fig:example_video}.
We conduct a comprehensive quantitative analysis and qualitative evaluation to verify the effectiveness of our \method, highlighting its superior performance compared to existing state-of-the-art methods. 

\section{Related work}

This work is highly relevant to the fields of video generation, human image animation, and temporal coherence modeling. We will give a brief discussion of them below.
\vspace{1mm}
\noindent
\textbf{Video generation.} 
Recent success in the diffusion models has remarkably boosted the progress of text-to-image generation~\cite{stablediffusion,GLIDE,Dalle2,T2i-adapter,huang2023composer,DDIM,DDPM,controlnet,saharia2022photorealistic,liu2023survey}.
However, generating videos from input conditions is a considerably more challenging task than its image counterpart due to the higher dimensional properties of video~\cite{make-a-video,imagenvideo}.
Different from static images, video exhibits an additional temporal dimension, comprising a sequence of frames, which is crucial for understanding dynamic visual content and textual input.
In order to model spatio-temporal dependencies, Make-A-Video~\cite{make-a-video} and ModelscopeT2V~\cite{modelscopet2v} adopt the 3D-UNet framework, which is a temporal extension of 2D-UNet~\cite{ronneberger2015unet} by integrating temporal layers such as temporal Transformers.
This paradigm has also been widely followed in subsequent work~\cite{tft2v,tune-a-video,chai2023stablevideo,ceylan2023pix2video,guo2023animatediff,zhou2022magicvideo,an2023latent,xing2023simda,qing2023hierarchical,yuan2023instructvideo,wei2023dreamvideo}.
In pursuit of higher controllability both spatially and temporally, some techniques such as Gen-1~\cite{Gen-1} and VideoComposer~\cite{videocomposer} attempt to introduce additional  guided conditions, \eg, depth maps and motion vectors, for controllable general video synthesis~\cite{zhao2023controlvideo,zhang2023controlvideo,xing2023make,yin2023dragnuwa,chen2023motion}.
In this work, we concentrate on the human-centered image animation task, which requires precise
control of both human-related appearance attributes and the desired target pose motion to create plausible videos.

\vspace{1mm}
\noindent
\textbf{Human image animation.} 
Animating human images along with the driving pose sequence is a challenging yet useful video creation task.
With the rapid development of generative networks, various approaches have been proposed. 
Previous works~\cite{FOMM,li2019dense,MRAA,TPS} mainly focus on exploring GAN-based generation, typically leveraging a motion network to predict dense appearance flows and perform feature warping on reference inputs to synthesize realistic images that follow the target poses.
However, these techniques often suffer from instability
training and mode collapse issues~\cite{Animateanyone}, struggling to precisely control the
generated human motions and yield sub-optimal synthesis quality.
As a result, extensive efforts~\cite{disco,Animateanyone,magicdance,champ,magicanimate,karras2023dreampose,xu2024you,zhu2024poseanimate} start to establish image animation architectures on diffusion models~\cite{DDIM,DDPM} due to its superior training stability and impressive high-fidelity results through iterative refining process.
For instance, Disco~\cite{disco} develops a hybrid diffusion architecture based on ControlNet~\cite{controlnet} with disentangled control of human foreground,
background, and pose to allow composable human dance generation.
MagicAnimate~\cite{magicanimate} and Animate Anyone~\cite{Animateanyone} take advantage of video diffusion
models for enhanced temporal consistency and introduce a two-stage learning strategy to decouple appearance alignment and motion guidance respectively.
Commonly,
existing diffusion-based frameworks usually build on a ControlNet-like 3D-UNet model~\cite{Animateanyone,guo2023animatediff,controlnet} to maintain temporal coherence and introduce a reference encoder, which is a replica of 3D-UNet excluding the temporal Transformer layers, to preserve the intricate appearance of the reference image.
Although promising, these advances often require multiple separate networks with non-negligible parameters, leading to increasing optimization difficulties, and face challenges in long-term video generation due to the quadratic complexity of the temporal Transformer.

\vspace{1mm}
\noindent
\textbf{Temporal coherence modeling.}
Temporal dynamics analysis plays a pivotal role in many video understanding tasks~\cite{lea2017temporal,wang2021oadtr,arnab2021vivit}.
Previous work on temporal modeling broadly falls into two categories: convolution-based~\cite{lea2017temporal,SSTAP,qiu2017learning} and RNN-based~\cite{mocogan,gupta2022rv,li2018video}.
However, convolution-based approaches suffer from limited receptive fields~\cite{bertasius2021space}, leading to difficulty in modeling long-range temporal dependencies.
Although RNN-based methods can perceive long-term relations, they encounter the dilemma of unparalleled calculation, causing computational inefficiency.
To capture non-local associations and enable parallel calculation, many recent researches~\cite{bertasius2021space,make-a-video,tune-a-video,videocomposer,wang2021oadtr,arnab2021vivit,liang2022transcrowd,li2023vigt,shao2021cpt,chen2023sparse,li2023monkey,liu2024textmonkey,wang2023few} seek to employ Transformers for sequential modeling, displaying remarkable performance in various downstream applications, \eg, action recognition~\cite{arnab2021vivit,bertasius2021space}, action detection~\cite{wang2021oadtr}, and video generation~\cite{guo2023animatediff,videocomposer,modelscopet2v}. Nevertheless, temporal Transformers still face huge computational costs due to the quadratic complexity, especially when dealing with long sequences.
Mamba~\cite{mamba}, a type of fundamental state space models (SSMs)~\cite{gu2021efficiently}, which conceptually merges the merits of parallelism and non-locality, has demonstrated convincing potential in a wide range of downstream natural language processing~\cite{mamba} and computer vision fields~\cite{Visionmamba,liu2024vmamba}.
Inspired by the excellent performance and linear time efficiency of Mamba  in long sequence processing~\cite{Visionmamba,yang2024plainmamba,liu2024vmamba,chen2024video,li2024videomamba}, this paper attempts to introduce Mamba into the human image animation task as a strong and promising alternative for temporal coherence modeling.

\section{UniAnimate}

Human image animation aims to generate a high-quality and temporally consistent video based on the input reference image and target pose sequence. The challenges in this task involve maintaining temporal consistency and natural appearance throughout the generated video. To this end, we present our proposed \method, which addresses the limitations of existing diffusion-based methods for consistent and long-term human image animation. We will first briefly introduce the basic concepts of the latent diffusion model. Subsequently, the detailed pipeline of \method will be described.

\subsection{Preliminaries of latent diffusion model}

The optimization and inference of traditional pixel-level diffusion models~\cite{Dalle2,DDIM,imagenvideo} requires prohibitive calculations in the high-dimensional
RGB image space. To reduce the computational cost, latent diffusion models~\cite{stablediffusion,wei2023dreamvideo,an2023latent,ma2023dreamtalk,zhang2023i2vgen} propose employing denoising procedures
in the latent space of a pre-trained variational autoencoder (VAE). 
In particular, a VAE encoder is first employed to embed the input sample to the down-sampled latent data $\mathbf{z}_0$.
Subsequently, a Markov chain of forward diffusion process $q$ is defined to progressively add stochastic Gaussian noise of $T$ steps to the clean latent data $\mathbf{z}_0$.
The forward diffusion step can be formulated as:
\begin{equation}
    q(\mathbf{z}_t \vert \mathbf{z}_{t-1}) = \mathcal{N}(\mathbf{z}_t; \sqrt{1 - \beta_t} \mathbf{z}_{t-1}, \beta_t\mathbf{I}), \quad t=1,2,...,T
\end{equation}
where $\beta_t \in (0, 1)$ denotes the noise schedule.  As $t$ gradually increases, the total noise imposed on the original $\mathbf{z}_0$ becomes more intense, and eventually $\mathbf{z}_t$ tends to be a random Gaussian noise.
The objective of the diffusion model $\boldsymbol{\epsilon}_{\theta}$ is to learn a reversed denoising process $p$ that aims to recover the desired clean sample $\mathbf{z}_0$  from the noised data $\mathbf{z}_t$.
The denoising process $p(\mathbf{z}_{t-1} \vert \mathbf{z}_t)$ can be estimated by $\boldsymbol{\epsilon}_{\theta}$ as the following form:
\begin{equation}
p_\theta(\mathbf{z}_{t-1} \vert \mathbf{z}_t) = \mathcal{N}(\mathbf{z}_{t-1}; \boldsymbol{\mu}_\theta(\mathbf{z}_t, t), \boldsymbol{\Sigma}_\theta(\mathbf{z}_t, t)) 
%
\end{equation}
where  $\boldsymbol{\mu}_\theta(\mathbf{z}_t, t)$ is the approximated objective of the reverse diffusion process, and $\theta$ means the parameters of the denoising model $\boldsymbol{\epsilon}_{\theta}$. 
In many video generation techniques~\cite{Animateanyone,videocomposer,VideoLDM}, the denoising model is a 3D-UNet model~\cite{VideoLDM}.
In the optimization stage, a simplified L2 loss is usually applied to minimize the discrepancies between predicted noise and real ground-truth noise:
\begin{equation}
\mathcal{L} = \mathbb{E}_{{\theta}} \Big[\|\boldsymbol{\epsilon} - \boldsymbol{\epsilon}_\theta(\mathbf{z}_t, t, c)\|^2 \Big]
\label{eq:objective}
\end{equation}
in which $c$ is the input conditional guidance. 
After the reversed denoising stage, the predicted clean latent is fed into the VAE decoder to reconstruct the predicted video in the pixel space.
%

\begin{figure}
    \centering
    \includegraphics[width=1.0\linewidth]{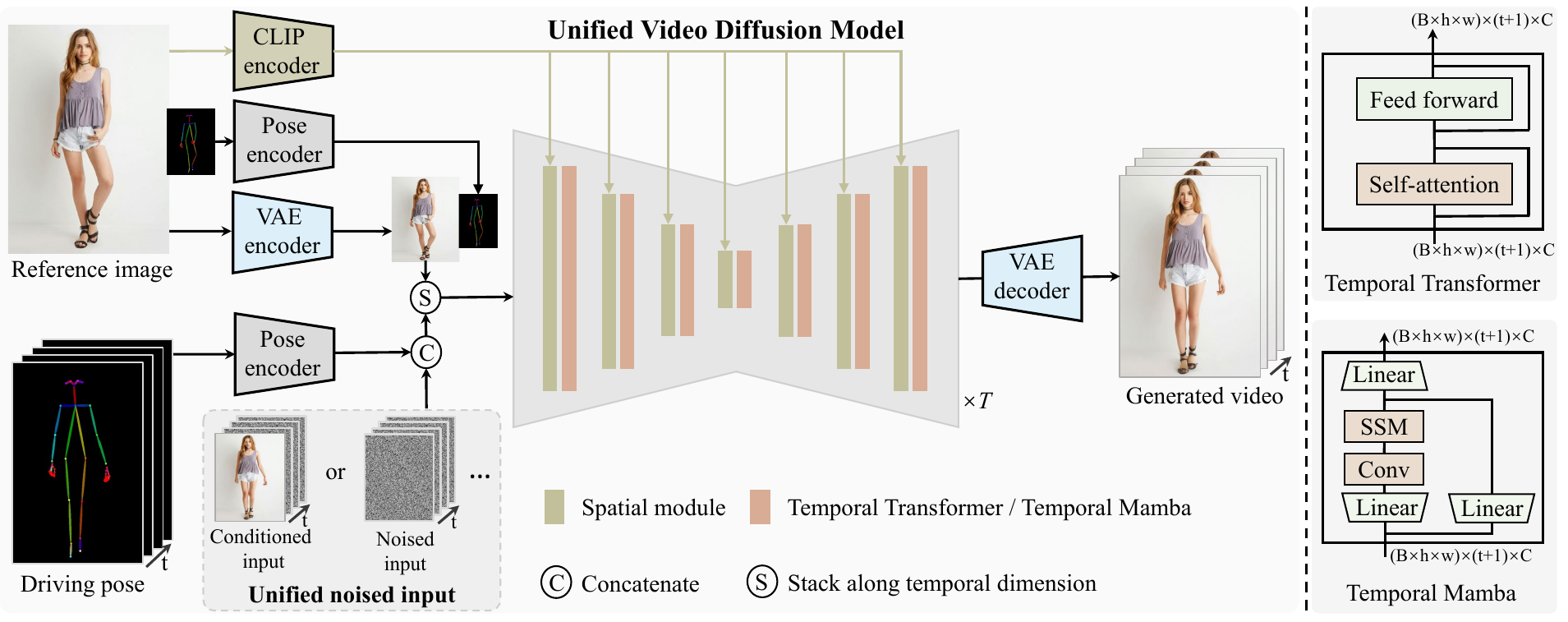}
    \vspace{-6mm}
    \caption{The overall architecture of the proposed \method. 
    Firstly, we utilize the CLIP encoder and VAE encoder to extract latent features of the given reference image. To facilitate the learning of the human body structure in the reference image, we also incorporate the representation of the reference pose into the final reference guidance.
Subsequently, we employ a pose encoder to encode the target driven pose sequence and concatenate it with the noised input along the channel dimension.
The noised input is derived from the first frame conditioned video or a noised video.
Then,
the concatenated noised input is stacked with the reference guidance along the temporal dimension and fed into the unified video diffusion model to remove noise. 
%
%
The temporal module in the unified video diffusion model can be the temporal Transformer or temporal Mamba.
Finally, a VAE decoder is adopted to map the generated latent video to the pixel space.
        }
    \label{fig:network}
\end{figure}

\subsection{{UniAnimate}}

\method aims to create visually appealing and temporally coherent videos that correspond to the given reference image and pose sequence. 
To align the appearance between the given image and the generated video, we design a unified video diffusion model to embed the reference information and estimated video content in the shared feature space.
%
In addition to the driving pose sequence, the source pose of the reference image is also incorporated to provide corresponding spatial position and layout information of the human body.
%
To ensure long-term video generation, a first frame conditioning strategy is introduced, and we explore an alternative based on Mamba~\cite{mamba} for temporal coherence modeling. The overall framework of the proposed \method is displayed in Figure~\ref{fig:network}.

\vspace{1mm}
\noindent
\textbf{Unified video diffusion model.}
To tackle the problem of temporally consistent human image animation, we leverage the widely used 3D-UNet structure~\cite{modelscopet2v,videocomposer,VideoLDM} for video creation. Unlike previous human image animation methods that employ two separate networks, namely a referenceNet for encoding the appearance of the reference image and a main 3D-UNet branch for synthesizing 
human motion videos, \method proposes to take advantage of a unified video diffusion model.
%
This unified structure is able to jointly encode the appearance of the reference image and synthesize the motion of the generated video. The advantages of this strategy are twofold: 1) the feature representations of the reference image and the generated video exist in the same feature space, facilitating appearance alignment, and 2) the parameters of the framework are reduced, making optimization more feasible. 
Additionally, different from previous methods~\cite{Animateanyone,magicanimate}, which need to learn character structure information implicitly from the reference image, we propose to extract the reference skeletal pose from the reference image, explicitly incorporating the position and layout information of the reference human.
Specifically,
the reference image is first encoded into latent space using a VAE encoder, resulting in a feature representation of size $C_{1} \times  h\times w$, where $C_{1}$, $h$, and $w$ represent the channel, width, and height respectively.
The reference pose is also processed through a pose encoder, extracting layout information.
The reference image and pose features are then fused to obtain the final reference representation $f_{ref}$ with a shape of $ C \times  h\times w$.
To incorporate target pose information, we use a pose encoder to encode the driving pose sequence and concatenate the resulting driving pose features and the input noised latent to obtain the fused features $f_{v} \in \mathbb{R}^{t \times C \times  h\times w}$, where $t$ means the temporal length.
Subsequently, the reference representation $f_{ref}$ and the fused features $f_{v}$ are stacked along the temporal dimension, resulting in combined features $f_{merge} \in \mathbb{R}^{(t+1) \times C \times  h\times w}$ .
Finally,
the combined features are then fed into the unified video diffusion model for jointly appearance alignment and motion modeling.

\vspace{1mm}
\noindent
\textbf{Unified noised input.}
Due to memory limitations, it is not possible to generate a long video in a single pass. Instead, multiple short video segments need to be synthesized separately and eventually merged into one long video. Typically, existing methods~\cite{magicanimate,Animateanyone} utilize the slide window strategy that employs temporally overlapped local windows to synthesize short videos and average the intersection parts
to generate longer videos.
However, in our experiments (Section~\ref{sec.Long_video_generation}), we empirically find that this  slide window strategy
may suffer from discontinuities between segments and usually can not preserve appearance consistency with the reference image. 
To address this issue, we propose a unified noised input that allows random noised video or first frame conditioned video as input for video synthesis.
The first frame conditioning manner takes the beginning frame of a video as the condition for generating videos starting from the frame. 
By leveraging this strategy, the last frame of the previous short video segment can be used as the first frame of the next segment, enabling seamless and visually coherent long-term animation. The first frame conditioning strategy offers two advantages: 1) it supports user-defined input images as the starting frame, combined with the target pose sequence for human image animation, and 2) 
it is able to generate consistent
long videos with smooth transitions by iteratively employing the first frame conditioning strategy.


\vspace{1mm}
\noindent
\textbf{Temporal modeling manners.}
Previous methods~\cite{videocomposer,VideoLDM,Animateanyone,magicanimate} usually employ temporal Transformers to model the motion patterns in the video. 
While these methods have shown impressive progress, the quadratic complexity relationship between temporal Transformers and input video length limits the video length that can be generated in a single segment. In this paper, we explore a new temporal modeling approach called temporal Mamba~\cite{mamba,Visionmamba,li2024videomamba} for the human image animation task. 
Mamba~\cite{mamba} is commonly treated as a type of linear time-invariant system that can map a sequential input $x(s) \in \mathbb{R}^{L}$ to a response state $y(s) \in \mathbb{R}^{L}$ and can be typically formulated as:
\begin{equation}
\begin{aligned}
{h'}(s) &= \mathbf{A}h(s) + \mathbf{B}x(s), \\
y(s) &= \mathbf{C}h(s)
\end{aligned}
\label{eq:mamba_1}
\end{equation}
where $\mathbf{A}$, $\mathbf{B}$ and $\mathbf{C}$ are parameter matrices, and $h(s)$ is a hidden state.
In the application, a discretized version of Mamba that adopts a bidirectional scanning mechanism~\cite{Visionmamba} is leveraged by us to handle temporal dependencies.
Temporal Mamba exhibits a linear complexity relationship with the generated video length.
As will be demonstrated in the experimental section, performance of temporal Mamba surpasses or matches that of temporal Transformers, while requiring less memory consumption.

\vspace{1mm}
\noindent
\textbf{Training and inference.}
During training, we follow the conventional video generation paradigm~\cite{VideoLDM,videocomposer} and train the model to generate clean videos by estimating the imposed noise. To facilitate the multi-condition generation, we introduce random dropout to the input conditions (\eg, the first frame and reference image) at a certain ratio (\eg, 0.5). 
At the inference stage, our \method supports human video animation using only a reference image and a target pose sequence, as well as the input of a first frame. 
To generate long videos composed of multiple segments, we utilize the reference image for the first segment. For subsequent segments, we use the reference image along with the first frame of the previous segment to initiate the next generation.

\section{Experiments}

In this section, we first describe the experimental setups of \method.
Afterward, a comprehensive qualitative and quantitative evaluation with existing state-of-the-art techniques will be implemented to validate the effectiveness of the proposed method in generating temporally smooth videos for the human image animation task.

%

\subsection{Experimental setups}

\noindent
\textbf{Datasets.}  Following previous works~\cite{Animateanyone,UBCfashion,TikTokdata}, the comparative experiments are conducted on two standard and widely-used datasets, namely TikTok~\cite{TikTokdata} and Fashion~\cite{UBCfashion}.
The TikTok dataset consists of 340 training videos and 100 testing videos. Each video has a duration of 10$\sim$15 seconds. To ensure a fair comparison, we follow the settings of prior methods~\cite{Animateanyone,disco,magicanimate}, where 10 videos from the test set are selected for both qualitative and quantitative comparisons.
Fashion is a dataset with simple and clean backgrounds, containing 500 training videos and 100 testing videos, with each video covering approximately 350 frames.
To enhance the robustness and generalization of our model, similar to~\cite{Animateanyone,champ}, we additionally collect around 10K TikTok-like internal videos. It is worth noting that, to enable fair comparisons with existing methods, we train our \method solely on the TikTok and Fashion benchmarks without incorporating extra videos and report experimental results in Section~\ref{Sec.Comparisons_SOTA} and Section~\ref{Sec.Ablation_study}.

\vspace{1mm}
\noindent
\textbf{Detailed implementation.} In the experiments, we use DWpose~\cite{DWpose} to extract pose sequences for model optimization.
The visual encoder of the multi-modal CLIP-Huge model~\cite{CLIP} in Stable Diffusion v2.1~\cite{stablediffusion} is used to encode CLIP embedding of the reference image.
The pose encoder is composed of several convolution layers and has a similar structure as STC-encoder in VideoComposer~\cite{videocomposer}.
Like previous approaches~\cite{magicanimate,Animateanyone,champ}, we employ a pre-trained video generation model~\cite{tft2v} for model initialization.
The experiments are conducted on 8$\sim$16 NVIDIA A100 GPUs. During the training phase, videos are resized to a spatial resolution of 768$\times$512. 
We randomly input video segments of uniformly sampled 16 or 32 frames into the model to learn temporal consistency.
We utilize the AdamW optimizer~\cite{loshchilov2017AdamW} with a learning rate of 5e-5 to optimize the network. 
For noise sampling, DDPM~\cite{DDPM} with 1000 steps is performed during training.
In the inference stage, we warp the length of the driving pose to roughly align with the reference pose and adopt the DDIM sampler~\cite{DDIM} with 50 steps for accelerated sampling.

\vspace{1mm}
\noindent
\textbf{Evaluation metrics.} We quantitatively evaluate our method using various metrics. 
In particular, four widely-used image metrics, namely L1, PSNR~\cite{hore2010image}, SSIM~\cite{wang2004image}, and LPIPS~\cite{zhang2018unreasonable}, are applied to measure the visual quality of the generated results. Besides these image metrics, we also leverage the Fréchet Video Distance (FVD)~\cite{unterthiner2018towards} as a video evaluation metric, which quantifies the discrepancy between the generated video distribution and the real video distribution.

\subsection{Comparisons with state-of-the-art methods}
\label{Sec.Comparisons_SOTA}

For a comprehensive evaluation,
we compare our proposed method with existing approaches in terms of both quantitative and qualitative measures.
Additionally, a human evaluation is further conducted to verify the efficacy.

\begin{table}[!t]
\caption{
Quantitative comparisons with existing methods on TikTok dataset.
``PSNR*'' indicates that the modified metric\protect\footnotemark{} is applied to avoid numerical overflow. Underline means the second best result.
}
%
\label{tab:quantitative_TikTok}
\renewcommand{\arraystretch}{1.15}
\setlength\tabcolsep{6.5pt}
\centering
\begin{tabular}{l|ccccc|c}
\shline
Method          & L1 $\downarrow$ & PSNR $\uparrow$ & PSNR* $\uparrow$ & SSIM $\uparrow$ & LPIPS $\downarrow$  & FVD $\downarrow$ \\ \shline
FOMM~\cite{FOMM} $_{\color{gray}{\text{(NeurIPS19)}}}$   & 3.61E-04        & -     & 17.26       & 0.648           & 0.335                                       & 405.22           \\
MRAA~\cite{MRAA} $_{\color{gray}{\text{(CVPR21)}}}$   & 3.21E-04        & -     & 18.14       & 0.672           & 0.296                                       & 284.82           \\
TPS~\cite{TPS} $_{\color{gray}{\text{(CVPR22)}}}$   & 3.23E-04        & -     & \underline{18.32}       & 0.673           & 0.299                                       & 306.17           \\
DreamPose~\cite{karras2023dreampose} $_{\color{gray}{\text{(ICCV23)}}}$  & 6.88E-04        & 28.11    & 12.82       & 0.511           & 0.442                            & 551.02           \\
DisCo~\cite{disco} $_{\color{gray}{\text{(CVPR24)}}}$            & 3.78E-04            & 29.03      & 16.55       & 0.668             & 0.292                                      & 292.80              \\
MagicAnimate~\cite{magicanimate} $_{\color{gray}{\text{(CVPR24)}}}$   & 3.13E-04    & 29.16  & -         & 0.714           & 0.239                                 & 179.07           \\
Animate Anyone~\cite{Animateanyone} $_{\color{gray}{\text{(CVPR24)}}}$   & -            & 29.56      & -       & 0.718             & 0.285                                            & 171.90              \\
Champ~\cite{champ} $_{\color{gray}{\text{(ArXiv24)}}}$ 
& \underline{2.94E-04}            & \underline{29.91}        & -     & \underline{0.802}            & \underline{0.234}                                        & \underline{160.82}         \\
\rowcolor{Gray}
\textbf{\method}  & \textbf{2.66E-04}       & \textbf{30.77}    & \textbf{20.58}             & \textbf{0.811}            & \textbf{0.231}                                          &\textbf{148.06}         \\

\shline
\end{tabular} 
\vspace{1mm}
\end{table}

\footnotetext{https://github.com/Wangt-CN/DisCo/issues/86\label{foot1}}

\begin{table}[!t]
\caption{
Quantitative comparisons with existing methods on the Fashion dataset.
``\emph{w/o Finetune}'' represents the method without additional finetuning on the fashion dataset.
``PSNR*'' indicates that the modified metric\protect\footref{foot1} is applied to avoid numerical overflow. Underline means the second best result.
}
\label{tab:quantitative_fashion}
\renewcommand{\arraystretch}{1.15}
\setlength\tabcolsep{8.5pt}
\centering
\begin{tabular}{l|cccc|c}
\shline
Method          & PSNR $\uparrow$ & PSNR* $\uparrow$ & SSIM $\uparrow$ & LPIPS $\downarrow$  & FVD $\downarrow$ \\ \shline

MRAA~\cite{MRAA} $_{\color{gray}{\text{(CVPR21)}}}$   &   -  &   - & 0.749 & 0.212 & 253.6   \\
TPS~\cite{TPS} $_{\color{gray}{\text{(CVPR22)}}}$   & - &   - & 0.746 & 0.213 & 247.5    \\
DPTN~\cite{DPTN} $_{\color{gray}{\text{(CVPR22)}}}$   & - &   24.00 & 0.907 & 0.060 & 215.1    \\
NTED~\cite{NTED} $_{\color{gray}{\text{(CVPR22)}}}$   & - &   22.03 & 0.890 & 0.073 & 278.9    \\
PIDM~\cite{bhunia2023person} $_{\color{gray}{\text{(CVPR23)}}}$   & - &   - & 0.713 & 0.288 & 1197.4    \\
DBMM~\cite{yu2023bidirectionally} $_{\color{gray}{\text{(ICCV23)}}}$   & -  &   \underline{24.07} & 0.918 & 0.048 & 168.3    \\
DreamPose~\cite{karras2023dreampose} $_{\color{gray}{\text{(ICCV23)}}}$  &  -  &   - & 0.885 & 0.068  & 238.7        \\
DreamPose \emph{w/o Finetune}~\cite{karras2023dreampose} $_{\color{gray}{\text{(ICCV23)}}}$  &  34.75 &   -  & 0.879 & 0.111 &  279.6        \\
Animate Anyone~\cite{Animateanyone} $_{\color{gray}{\text{(CVPR24)}}}$   &  \textbf{{38.49}} &   - & \underline{0.931} & \underline{0.044} & \underline{81.6}             \\
\rowcolor{Gray}
\textbf{\method}  & \underline{37.92}          &   \textbf{27.56}        & \textbf{0.940}            & \textbf{0.031}                                            & \textbf{68.1}         \\

\shline
\end{tabular} 
\vspace{1mm}
\end{table}

\begin{figure}[!t]
    \centering
    \includegraphics[width=0.98\linewidth]{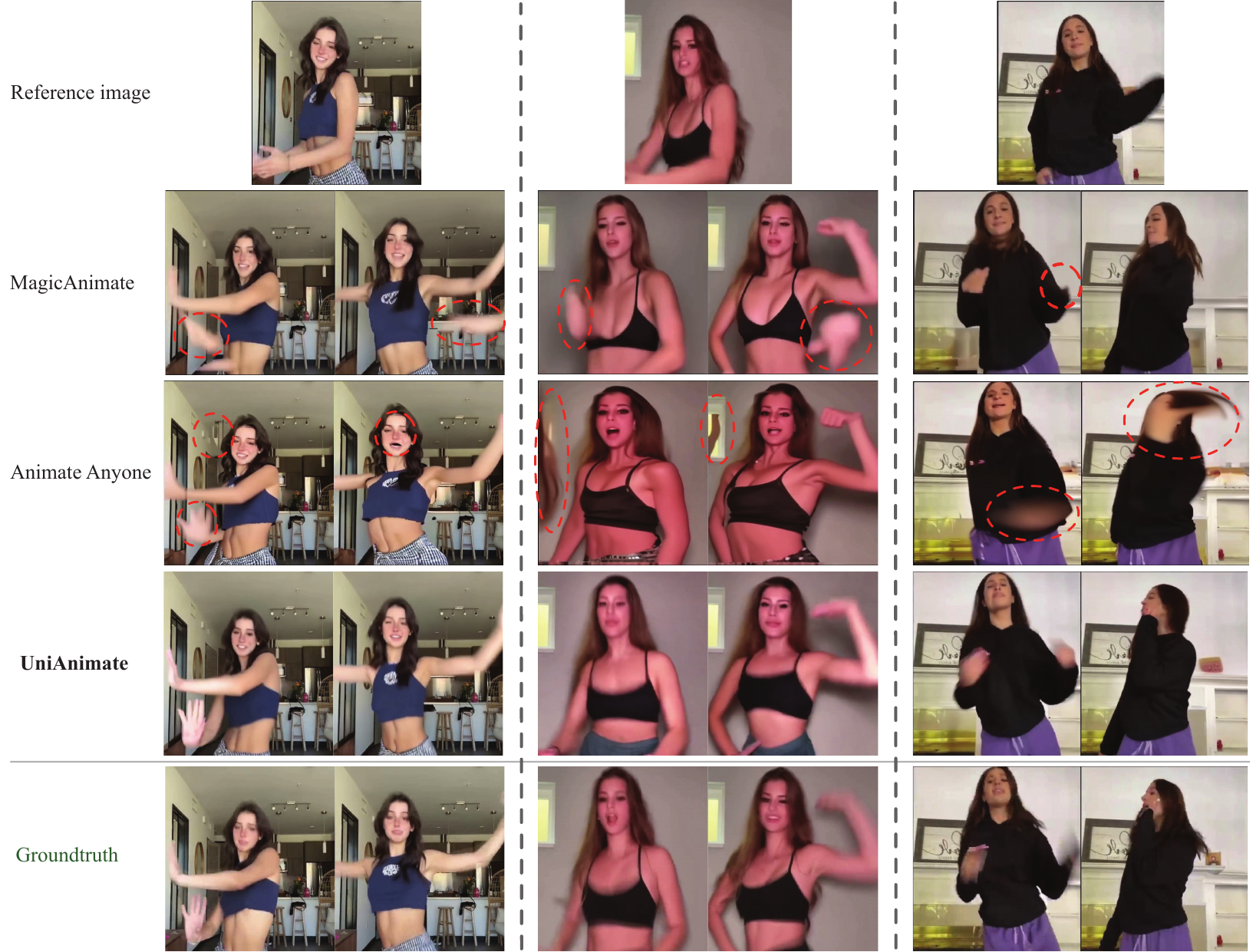}
    \vspace{-2mm}
    \caption{Qualitative comparison with existing state-of-the-art methods on the TikTok dataset. Two typical state-of-the-art methods namely MagicAnimate~\cite{magicanimate} and Anymate Anyone~\cite{Animateanyone} are compared.
        }
    \label{fig:compare_with_sota}
\end{figure}

\vspace{1mm}
\noindent
\textbf{Quantitative comparisons.}  To validate the effectiveness of our proposed method, we compare it with existing state-of-the-art approaches, including Disco~\cite{disco}, MagicAnimate~\cite{magicanimate}, Animate Anyone~\cite{Animateanyone}, Champ~\cite{champ}, \etc. These methods adopt ControlNet-like structures to achieve appearance alignment. 
As shown in Table~\ref{tab:quantitative_TikTok}, our \method outperforms existing state-of-the-art competitors across all the evaluation metrics on the TikTok dataset.
For example, 
\method reaches a FVD of 148.06, achieving the best video fidelity among recent works.
The quantitative results of both image and video metrics demonstrate our model's excellent ability to learn and generate realistic content, highlighting the capability of \method to effectively capture and reproduce the underlying data distribution of training samples.
The experiment on the Fashion dataset is also conducted, as illustrated in Table~\ref{tab:quantitative_fashion}.
From the comparison, we can observe that \method exhibits superior structural preservation capacity, obtaining the best SSIM of 0.940.
\method also achieves impressive performance on other metrics, and these results collectively reaffirm the ability of \method to synthesize visually fidelity animations in the fashion video domain.
%

\vspace{1mm}
\noindent
\textbf{Qualitative comparisons.} 
In addition to quantitative measures, we also provide a qualitative comparison in Figure~\ref{fig:compare_with_sota}. We showcase the comparison of \method with other competitive methods on the TikTok test set. The results of Animate Anyone~\cite{Animateanyone} are obtained by leveraging the publicly available reproduced code\footnote{https://github.com/MooreThreads/Moore-AnimateAnyone}. 
From the visualizations, we can observe that MagicAnimate~\cite{magicanimate} exhibits instances of limb generation and appearance misalignment, while Animate Anyone introduces undesirable artifacts. These methods fail to produce satisfactory results. In contrast, the proposed \method consistently generates high-quality and coherent pose transfer results that adhere to the input conditions, demonstrating remarkable controllability.
We attribute our advanced performance to the use of a unified video diffusion model to handle both reference image and noised video simultaneously, resulting in a common feature space for appearance alignment and motion modeling, facilitating model optimization.

\vspace{1mm}
\noindent
\textbf{Human evaluation.}  In order to further assess the performance of our method, we incorporate an additional human evaluation, as depicted in Table~\ref{tab:user_study}. 
We asked users to rate the generated results with respect to visual quality, identity preservation, and temporal consistency. 
The human evaluation results indicate that our method displays
favorable visual aesthetics, reliable controllability, and enhanced temporal consistency. 

\begin{table}[!t]
\caption{
User study. We ask users to rate the generated video results on the TikTok dataset in terms of visual quality, identify preservation, and temporal consistency.
}
\label{tab:user_study}
\renewcommand{\arraystretch}{1.15}
\setlength\tabcolsep{5pt}
\centering
\begin{tabular}{l|ccc}
\shline
Method          & Visual quality $\uparrow$ & Identity preservation $\uparrow$ & Temporal consistency $\uparrow$  \\ \shline
MagicAnimate~\cite{magicanimate} 
& 76\% & 81\% & 82\%        \\
Animate Anyone~\cite{Animateanyone} 
& 67\% & 84\% & 71\%        \\
Champ~\cite{champ} 
 & 74\% & 77\% & 85\%        \\
\rowcolor{Gray}
\textbf{\method}   & \textbf{85\%} & \textbf{89\%} & \textbf{91\%}              \\

\shline
\end{tabular} 
\vspace{1mm}
\end{table}

\begin{table}[!t]
\caption{
Ablation study on the TikTok dataset.
``PSNR*'' indicates that the modified metric is applied to avoid numerical overflow. 
``\method \emph{w/o} unified VDM'' implies a ControlNet-like structure, \ie, two separate networks to encode the appearance and temporal coherence respectively.
%
}
%
\label{tab:ablation_study}
\renewcommand{\arraystretch}{1.15}
\setlength\tabcolsep{6.5pt}
\centering
\begin{tabular}{l|cccc|c}
\shline
Method          & L1 $\downarrow$ &    PSNR* $\uparrow$ & SSIM $\uparrow$ & LPIPS $\downarrow$  & FVD $\downarrow$ \\ \shline

\method \emph{w/o} reference pose  & 3.07E-04            & 18.45            & 0.735           & 0.276                                         & 182.41        \\
\method \emph{w/o} unified VDM  &  3.12E-04           & 18.09             & 0.712           & 0.291                                         & 205.28        \\
\rowcolor{Gray}
\textbf{\method}  & \textbf{2.66E-04}         & \textbf{20.58}             & \textbf{0.811}            & \textbf{0.231}                                          &\textbf{148.06}         \\

\shline
\end{tabular} 
\vspace{1mm}
\end{table}

\begin{figure}[!t]
    \centering
    \includegraphics[width=0.98\linewidth]{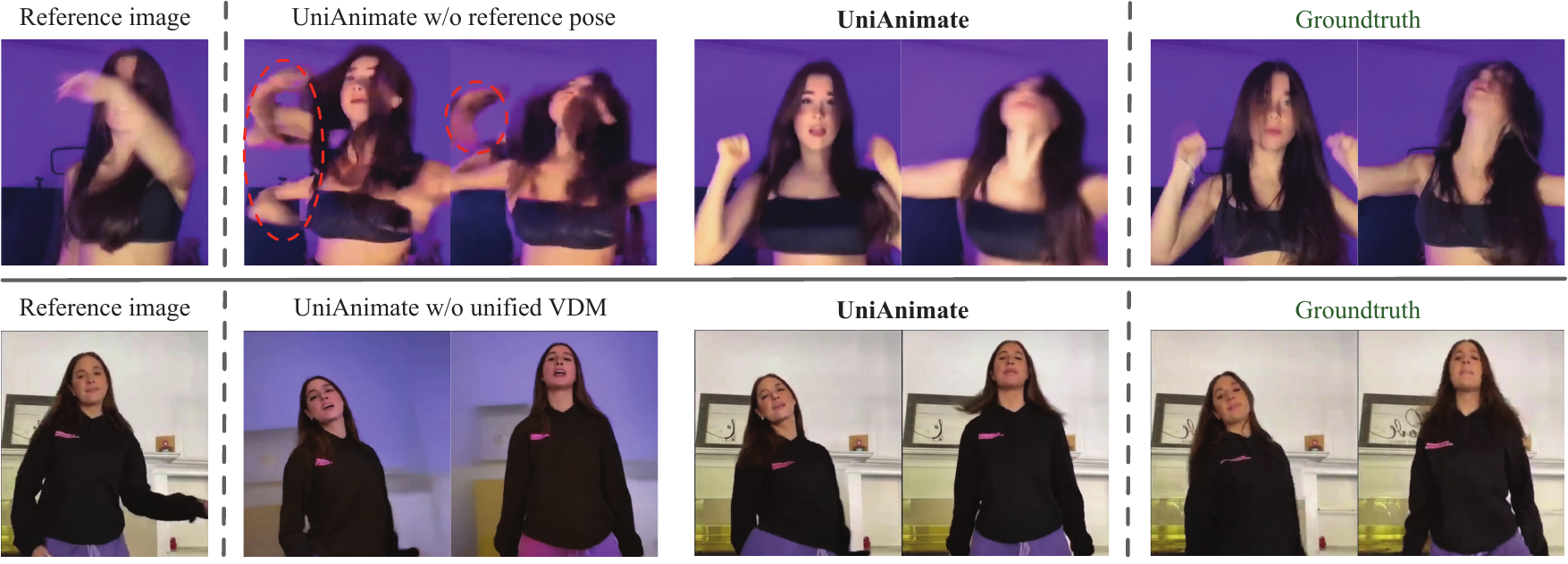}
    \vspace{-2mm}
    \caption{Ablation study. To ensure a fair comparison, the same random noise is imposed on the baseline methods and  \method.
        }
    \label{fig:ablation_study}
\end{figure}

\subsection{Ablation study}
\label{Sec.Ablation_study}


\noindent
\textbf{Analysis of network components.}
To generate temporally consistent videos that are visually aligned with the given reference image, we introduce a unified video diffusion model. 
This architecture employs a shared 3D-UNet  to handle both appearance alignment and motion modeling.
To further improve the appearance alignment, we
incorporate a reference pose to facilitate understanding of the reference image's layout and human structure. 
We conduct an ablation study on the proposed unified video diffusion model architecture and the effect of reference pose, as shown in Table~\ref{tab:ablation_study}. The results indicate that each module contributes significantly to the overall performance improvement. 
Example cases in Figure~\ref{fig:ablation_study} further demonstrate the crucial roles of each module. 
For instance,
removing the reference pose may result in undesired artifacts such as ``disconnected limbs.'' On the other hand, the synthesized results may display appearance inconsistencies (\eg, mismatched backgrounds) with the reference image without the unified video diffusion model.  
This can be attribute to the fact that aligning features into the same space becomes a challenging task with two separate networks. 
In contrast,
our method exhibits remarkable results in appearance alignment.

\begin{table}[!t]
\caption{
Quantitative comparison of different temporal modeling manners on the TikTok dataset.
}
%
\label{tab:quantitative_TikTok_temporal_manner}
\renewcommand{\arraystretch}{1.15}
\setlength\tabcolsep{6.5pt}
\centering
\begin{tabular}{l|cccc|c}
\shline
Method          & L1 $\downarrow$ &    PSNR* $\uparrow$ & SSIM $\uparrow$ & LPIPS $\downarrow$  & FVD $\downarrow$ \\ \shline

Temporal Mamba  & \textbf{2.47E-04}            & \textbf{20.81}            & 0.804          & \textbf{0.222}                                         & 156.26        \\

Temporal Transformer (Default)  & {2.66E-04}         & {20.58}             & \textbf{{0.811}}            & {0.231}                                          & \textbf{{148.06}}         \\

\shline
\end{tabular} 
\vspace{1mm}
\end{table}

\begin{table}[!t]
\caption{
Quantitative comparison of different temporal modeling manners about inference memory cost (GB).
``OOM'' is short for out of memory. Experiments are conducted on NVIDIA 80G A100 GPUs.
Note that inference memory and training memory are not the same, and training memory will be much larger since extra gradient calculations are involved. 
The inference overhead is for the entire framework, including CLIP encoder, pose encoder, \etc.
%
}
%
\label{tab:quantitative_memory_temporal_manner}
\renewcommand{\arraystretch}{1.15}
\setlength\tabcolsep{5.6pt}
\centering
\begin{tabular}{l|ccccc}
\shline
Method          & 32 Frames &    64 Frames & 128 Frames & 256 Frames & 360 Frames   \\ \shline

Temporal Mamba  & 24.1G            & 28.6G            & 43.3G          & 62.4G    &  78.8G                                           \\
Temporal Transformer  &   25.9G       & 32.9G            & 50.6G            & 73.6G   & OOM                                            \\

\shline
\end{tabular} 
\vspace{1mm}
\end{table}

\begin{figure}[!t]
    \centering
    \includegraphics[width=0.98\linewidth]{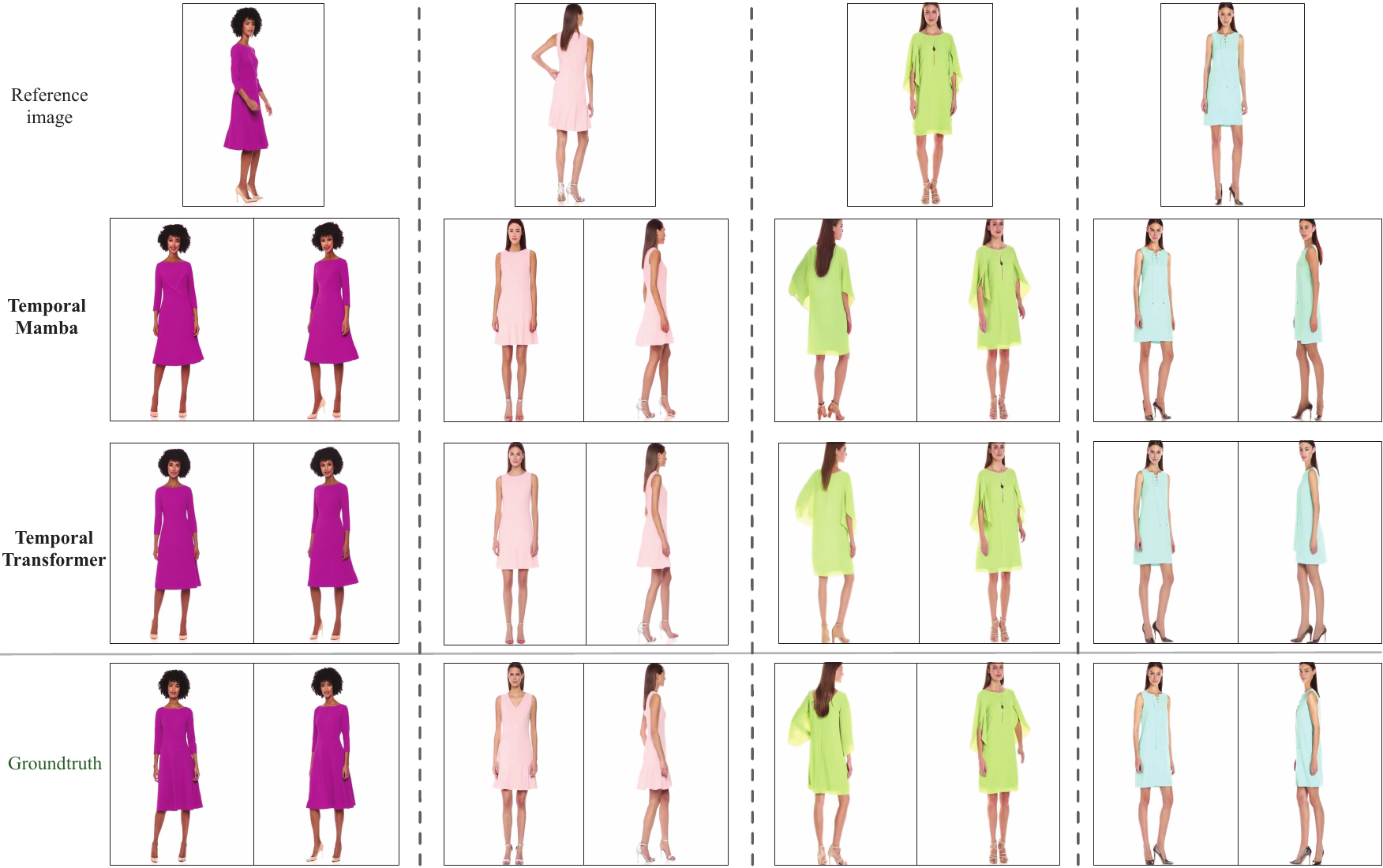}
    \vspace{-2mm}
    \caption{Generated video examples on the Fashion dataset. Results are generated by \method with temporal Mamba and temporal Transformer.
        }
    \label{fig:Mamba_results}
\end{figure}

\begin{figure}[!t]
    \centering
    \includegraphics[width=0.98\linewidth]{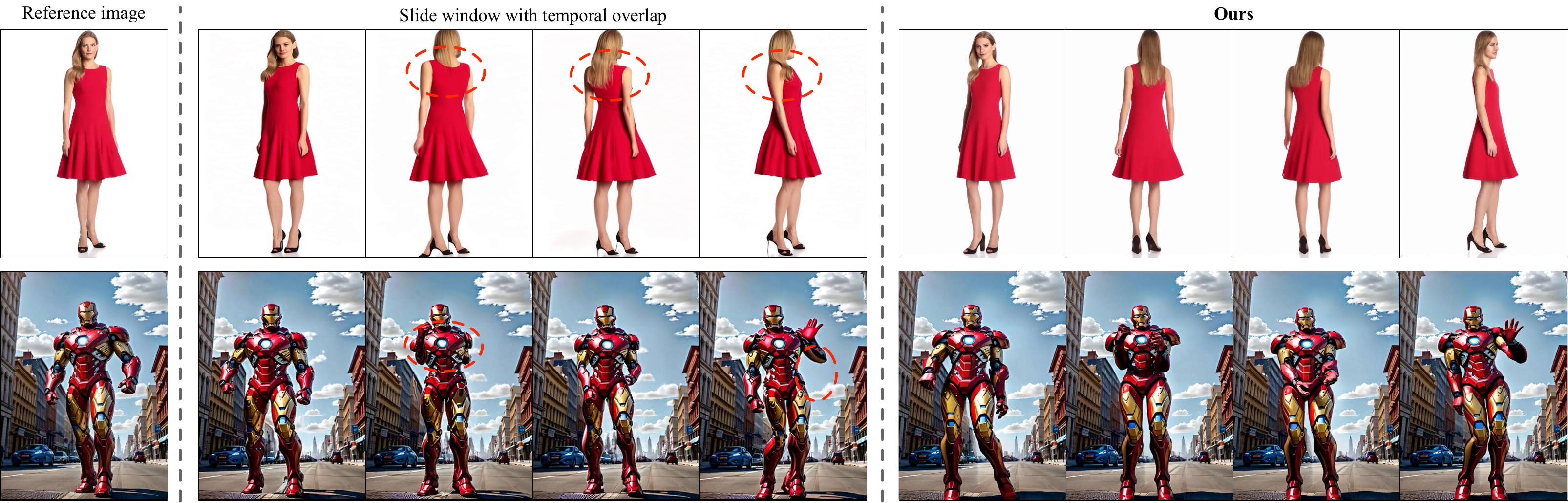}
    \vspace{-2mm}
    \caption{
    Qualitative comparison of different long video generation strategies.
    Existing method use the slide window strategy may straggle to synthesize smooth transitions, resulting in discontinuous appearance and inconsistent background.
    %
        %
        }
    \label{fig:transition}
\end{figure}

\vspace{1mm}
\noindent
\textbf{Varying temporal modeling manners.}
In our \method, we introduce the temporal Mamba as an alternative component for temporal modeling in the human image animation model. As illustrated in Table~\ref{tab:quantitative_TikTok_temporal_manner}, we find that temporal Mamba achieves comparable performance to temporal Transformer, both providing effective temporal modeling. 
From the comparative results in Figure~\ref{fig:Mamba_results}, we notice that both temporal Mamba and temporal Transformer exhibit excellent visually appealing results under our \method framework, and the generated results are basically close to real videos without any obvious artifacts.
Notably, we observe experimentally that temporal Mamba is particularly suitable for handling long sequences in terms of memory cost, as the computational resource overhead grows linearly with time, as shown in Table~\ref{tab:quantitative_memory_temporal_manner}. We hope that this proposed temporal modeling mechanism can lay the foundation for future research in this domain, especially long-range temporal modeling.



\section{Long video generation with smooth transitions}
\label{sec.Long_video_generation}

To enable the generation of long videos, we incorporate the unified noised input that supports first-frame conditioning, which allows us to continue generating subsequent video frames by leveraging the final frame of the previously generated segment and the reference image. 
As illustrated in Figure~\ref{fig:transition}, we compare our first-frame conditioning solution with the slide window strategy used in~\cite{magicanimate} and observe that the slide window strategy may suffer from unsatisfactory transition results, such as discontinuous appearance and inconsistent background.
%
We attribute this to the fact that the input pose sequence and the difficulty of denoising are different between two adjacent windows, so directly averaging the intersecting parts may damage the generated results and bring in artifacts.
In contrast, the first-frame conditioning technique used in our \method can keep the last frame of the previous segment the same as the beginning frame of the following segment, thus achieving a smooth transition. 

%

\section{Limitations}

Although \method achieves superior results compared to existing state-of-the-art approaches, there are still some limitations: \emph{i)} 
Generating realistic and fine-grained details in facial and hand regions remains challenging.
%
\emph{ii)} 
During the long video generation process, the completion of invisible parts by different video segments may be inconsistent.
This inconsistency can occasionally lead to temporal artifacts, disrupting the overall continuity of the generated videos.
In the future, we will focus on collecting high-quality HD videos and designing cross-segment interaction strategies to achieve more consistent human image animation results.

\section{Conclusion}

In this paper, we presented \method, a novel approach for generating high-fidelity, temporally smooth videos for human image animation. By introducing the unified video diffusion model, the unified noised input, and temporal Mamba, we address the appearance misalignment limitation  of existing methods and achieve improved video generation quality and efficiency. Extensive experimental results quantitatively and qualitatively validate the effectiveness of the proposed \method and highlight its potential for practical application deployment.

\Acknowledgements{This work is supported by the National Natural Science Foundation
of China under grant U22B2053 and 623B2039, and Alibaba Group through Alibaba Research Intern Program.}


{
\bibliography{egbib.bib}

\begin{thebibliography}{10}
\providecommand{\bibAnnoteFile}[1]{%
  \IfFileExists{#1}{\begin{quotation}\noindent\textsc{Key:} #1\\
  \textsc{Annotation:}\ \input{#1}\end{quotation}}{}}
\providecommand{\bibAnnote}[2]{%
  \begin{quotation}\noindent\textsc{Key:} #1\\
  \textsc{Annotation:}\ #2\end{quotation}}

\bibitem{an2023latent}
An J, Zhang S, Yang H, et~al.
\newblock Latent-shift: Latent diffusion with temporal shift for efficient text-to-video generation.
\newblock arXiv preprint arXiv:230408477, 2023
\bibAnnoteFile{an2023latent}

\bibitem{arnab2021vivit}
Arnab A, Dehghani M, Heigold G, et~al.
\newblock Vivit: A video vision transformer.
\newblock In: ICCV. 2021, 6836--6846
\bibAnnoteFile{arnab2021vivit}

\bibitem{bertasius2021space}
Bertasius G, Wang H, and Torresani L.
\newblock Is space-time attention all you need for video understanding?
\newblock In: ICML. 2021, volume~2, 4
\bibAnnoteFile{bertasius2021space}

\bibitem{bhunia2023person}
Bhunia A~K, Khan S, Cholakkal H, et~al.
\newblock Person image synthesis via denoising diffusion model.
\newblock In: CVPR. 2023, 5968--5976
\bibAnnoteFile{bhunia2023person}

\bibitem{VideoLDM}
Blattmann A, Rombach R, Ling H, et~al.
\newblock Align your latents: High-resolution video synthesis with latent diffusion models.
\newblock In: CVPR. 2023, 22563--22575
\bibAnnoteFile{VideoLDM}

\bibitem{ceylan2023pix2video}
Ceylan D, Huang C~H~P, and Mitra N~J.
\newblock Pix2video: Video editing using image diffusion.
\newblock In: ICCV. 2023, 23206--23217
\bibAnnoteFile{ceylan2023pix2video}

\bibitem{chai2023stablevideo}
Chai W, Guo X, Wang G, et~al.
\newblock Stablevideo: Text-driven consistency-aware diffusion video editing.
\newblock In: ICCV. 2023, 23040--23050
\bibAnnoteFile{chai2023stablevideo}

\bibitem{magicdance}
Chang D, Shi Y, Gao Q, et~al.
\newblock Magicdance: Realistic human dance video generation with motions \& facial expressions transfer.
\newblock arXiv preprint arXiv:231112052, 2023
\bibAnnoteFile{magicdance}

\bibitem{chen2024video}
Chen G, Huang Y, Xu J, et~al.
\newblock Video mamba suite: State space model as a versatile alternative for video understanding.
\newblock arXiv preprint arXiv:240309626, 2024
\bibAnnoteFile{chen2024video}

\bibitem{chen2023sparse}
Chen H, Li H, Li Y, et~al.
\newblock Sparse spatial transformers for few-shot learning.
\newblock Science China Information Sciences, 2023.
\newblock 66:210102
\bibAnnoteFile{chen2023sparse}

\bibitem{chen2023videocrafter1}
Chen H, Xia M, He Y, et~al.
\newblock Videocrafter1: Open diffusion models for high-quality video generation.
\newblock arXiv preprint arXiv:231019512, 2023
\bibAnnoteFile{chen2023videocrafter1}

\bibitem{chen2023motion}
Chen T~S, Lin C~H, Tseng H~Y, et~al.
\newblock Motion-conditioned diffusion model for controllable video synthesis.
\newblock arXiv preprint arXiv:230414404, 2023
\bibAnnoteFile{chen2023motion}

\bibitem{Gen-1}
Esser P, Chiu J, Atighehchian P, et~al.
\newblock Structure and content-guided video synthesis with diffusion models.
\newblock In: ICCV. 2023, 7346--7356
\bibAnnoteFile{Gen-1}

\bibitem{goodfellow2014generative}
Goodfellow I, Pouget-Abadie J, Mirza M, et~al.
\newblock Generative adversarial nets.
\newblock NeurIPS, 2014.
\newblock 27
\bibAnnoteFile{goodfellow2014generative}

\bibitem{mamba}
Gu A and Dao T.
\newblock Mamba: Linear-time sequence modeling with selective state spaces.
\newblock arXiv preprint arXiv:231200752, 2023
\bibAnnoteFile{mamba}

\bibitem{gu2021efficiently}
Gu A, Goel K, and R{\'e} C.
\newblock Efficiently modeling long sequences with structured state spaces.
\newblock arXiv preprint arXiv:211100396, 2021
\bibAnnoteFile{gu2021efficiently}

\bibitem{guo2023animatediff}
Guo Y, Yang C, Rao A, et~al.
\newblock Animatediff: Animate your personalized text-to-image diffusion models without specific tuning.
\newblock arXiv preprint arXiv:230704725, 2023
\bibAnnoteFile{guo2023animatediff}

\bibitem{gupta2022rv}
Gupta S, Keshari A, and Das S.
\newblock Rv-gan: Recurrent gan for unconditional video generation.
\newblock In: CVPR. 2022, 2024--2033
\bibAnnoteFile{gupta2022rv}

\bibitem{imagenvideo}
Ho J, Chan W, Saharia C, et~al.
\newblock Imagen video: High definition video generation with diffusion models.
\newblock arXiv preprint arXiv:221002303, 2022
\bibAnnoteFile{imagenvideo}

\bibitem{DDPM}
Ho J, Jain A, and Abbeel P.
\newblock Denoising diffusion probabilistic models.
\newblock NeurIPS, 2020.
\newblock 33:6840--6851
\bibAnnoteFile{DDPM}

\bibitem{cogvideo}
Hong W, Ding M, Zheng W, et~al.
\newblock Cogvideo: Large-scale pretraining for text-to-video generation via {T}ransformers.
\newblock In: ICLR. 2023
\bibAnnoteFile{cogvideo}

\bibitem{hore2010image}
Hore A and Ziou D.
\newblock Image quality metrics: Psnr vs. ssim.
\newblock In: ICPR. 2010, 2366--2369
\bibAnnoteFile{hore2010image}

\bibitem{Animateanyone}
Hu L, Gao X, Zhang P, et~al.
\newblock Animate anyone: Consistent and controllable image-to-video synthesis for character animation.
\newblock arXiv preprint arXiv:231117117, 2023
\bibAnnoteFile{Animateanyone}

\bibitem{huang2023composer}
Huang L, Chen D, Liu Y, et~al.
\newblock Composer: Creative and controllable image synthesis with composable conditions.
\newblock ICML, 2023
\bibAnnoteFile{huang2023composer}

\bibitem{TikTokdata}
Jafarian Y and Park H~S.
\newblock Learning high fidelity depths of dressed humans by watching social media dance videos.
\newblock In: CVPR. 2021, 12753--12762
\bibAnnoteFile{TikTokdata}

\bibitem{jiang2022text2human}
Jiang Y, Yang S, Qiu H, et~al.
\newblock Text2human: Text-driven controllable human image generation.
\newblock ACM Transactions on Graphics, 2022.
\newblock 41:1--11
\bibAnnoteFile{jiang2022text2human}

\bibitem{karras2023dreampose}
Karras J, Holynski A, Wang T~C, et~al.
\newblock Dreampose: Fashion video synthesis with stable diffusion.
\newblock In: ICCV. 2023, 22680--22690
\bibAnnoteFile{karras2023dreampose}

\bibitem{lea2017temporal}
Lea C, Flynn M~D, Vidal R, et~al.
\newblock Temporal convolutional networks for action segmentation and detection.
\newblock In: CVPR. 2017, 156--165
\bibAnnoteFile{lea2017temporal}

\bibitem{li2023vigt}
Li K, Guo D, and Wang M.
\newblock Vigt: proposal-free video grounding with a learnable token in the transformer.
\newblock Science China Information Sciences, 2023.
\newblock 66:202102
\bibAnnoteFile{li2023vigt}

\bibitem{li2024videomamba}
Li K, Li X, Wang Y, et~al.
\newblock Videomamba: State space model for efficient video understanding.
\newblock arXiv preprint arXiv:240306977, 2024
\bibAnnoteFile{li2024videomamba}

\bibitem{li2019dense}
Li Y, Huang C, and Loy C~C.
\newblock Dense intrinsic appearance flow for human pose transfer.
\newblock In: CVPR. 2019, 3693--3702
\bibAnnoteFile{li2019dense}

\bibitem{li2018video}
Li Y, Min M, Shen D, et~al.
\newblock Video generation from text.
\newblock In: AAAI. 2018, volume~32
\bibAnnoteFile{li2018video}

\bibitem{li2023monkey}
Li Z, Yang B, Liu Q, et~al.
\newblock Monkey: Image resolution and text label are important things for large multi-modal models.
\newblock In: CVPR. 2024
\bibAnnoteFile{li2023monkey}

\bibitem{liang2022transcrowd}
Liang D, Chen X, Xu W, et~al.
\newblock Transcrowd: weakly-supervised crowd counting with transformers.
\newblock Science China Information Sciences, 2022.
\newblock 65:160104
\bibAnnoteFile{liang2022transcrowd}

\bibitem{liu2023survey}
Liu M, Wei Y, Wu X, et~al.
\newblock Survey on leveraging pre-trained generative adversarial networks for image editing and restoration.
\newblock Science China Information Sciences, 2023.
\newblock 66:151101
\bibAnnoteFile{liu2023survey}

\bibitem{liu2024vmamba}
Liu Y, Tian Y, Zhao Y, et~al.
\newblock Vmamba: Visual state space model.
\newblock arXiv preprint arXiv:240110166, 2024
\bibAnnoteFile{liu2024vmamba}

\bibitem{liu2024textmonkey}
Liu Y, Yang B, Liu Q, et~al.
\newblock Textmonkey: An ocr-free large multimodal model for understanding document.
\newblock arXiv preprint arXiv:240304473, 2024
\bibAnnoteFile{liu2024textmonkey}

\bibitem{loshchilov2017AdamW}
Loshchilov I and Hutter F.
\newblock Decoupled weight decay regularization.
\newblock arXiv preprint arXiv:171105101, 2017
\bibAnnoteFile{loshchilov2017AdamW}

\bibitem{luo2023videofusion}
Luo Z, Chen D, Zhang Y, et~al.
\newblock Videofusion: Decomposed diffusion models for high-quality video generation.
\newblock In: CVPR. 2023, 10209--10218
\bibAnnoteFile{luo2023videofusion}

\bibitem{ma2024follow}
Ma Y, He Y, Cun X, et~al.
\newblock Follow your pose: Pose-guided text-to-video generation using pose-free videos.
\newblock In: Proceedings of the AAAI Conference on Artificial Intelligence. 2024, volume~38, 4117--4125
\bibAnnoteFile{ma2024follow}

\bibitem{ma2023dreamtalk}
Ma Y, Zhang S, Wang J, et~al.
\newblock Dreamtalk: When expressive talking head generation meets diffusion probabilistic models.
\newblock arXiv preprint arXiv:231209767, 2023
\bibAnnoteFile{ma2023dreamtalk}

\bibitem{T2i-adapter}
Mou C, Wang X, Xie L, et~al.
\newblock T2i-adapter: Learning adapters to dig out more controllable ability for text-to-image diffusion models.
\newblock arXiv preprint arXiv:230208453, 2023
\bibAnnoteFile{T2i-adapter}

\bibitem{GLIDE}
Nichol A~Q, Dhariwal P, Ramesh A, et~al.
\newblock Glide: Towards photorealistic image generation and editing with text-guided diffusion models.
\newblock In: ICML. 2022, 16784--16804
\bibAnnoteFile{GLIDE}

\bibitem{qing2023hierarchical}
Qing Z, Zhang S, Wang J, et~al.
\newblock Hierarchical spatio-temporal decoupling for text-to-video generation.
\newblock arXiv preprint arXiv:231204483, 2023
\bibAnnoteFile{qing2023hierarchical}

\bibitem{qiu2017learning}
Qiu Z, Yao T, and Mei T.
\newblock Learning spatio-temporal representation with pseudo-3d residual networks.
\newblock In: ICCV. 2017, 5533--5541
\bibAnnoteFile{qiu2017learning}

\bibitem{CLIP}
Radford A, Kim J~W, Hallacy C, et~al.
\newblock Learning transferable visual models from natural language supervision.
\newblock In: ICML. 2021, 8748--8763
\bibAnnoteFile{CLIP}

\bibitem{Dalle2}
Ramesh A, Dhariwal P, Nichol A, et~al.
\newblock Hierarchical text-conditional image generation with clip latents.
\newblock arXiv preprint arXiv:220406125, 2022.
\newblock 1:3
\bibAnnoteFile{Dalle2}

\bibitem{NTED}
Ren Y, Fan X, Li G, et~al.
\newblock Neural texture extraction and distribution for controllable person image synthesis.
\newblock In: CVPR. 2022, 13535--13544
\bibAnnoteFile{NTED}

\bibitem{stablediffusion}
Rombach R, Blattmann A, Lorenz D, et~al.
\newblock High-resolution image synthesis with latent diffusion models.
\newblock In: CVPR. 2022, 10684--10695
\bibAnnoteFile{stablediffusion}

\bibitem{ronneberger2015unet}
Ronneberger O, Fischer P, and Brox T.
\newblock U-net: Convolutional networks for biomedical image segmentation.
\newblock In: MICCAI. 2015, 234--241
\bibAnnoteFile{ronneberger2015unet}

\bibitem{saharia2022photorealistic}
Saharia C, Chan W, Saxena S, et~al.
\newblock Photorealistic text-to-image diffusion models with deep language understanding.
\newblock NeurIPS, 2022.
\newblock 35:36479--36494
\bibAnnoteFile{saharia2022photorealistic}

\bibitem{shao2021cpt}
Shao Y, Geng Z, Liu Y, et~al.
\newblock Cpt: A pre-trained unbalanced transformer for both chinese language understanding and generation.
\newblock Science China Information Sciences, 2024
\bibAnnoteFile{shao2021cpt}

\bibitem{FOMM}
Siarohin A, Lathuili{\`e}re S, Tulyakov S, et~al.
\newblock First order motion model for image animation.
\newblock NeurIPS, 2019.
\newblock 32
\bibAnnoteFile{FOMM}

\bibitem{MRAA}
Siarohin A, Woodford O~J, Ren J, et~al.
\newblock Motion representations for articulated animation.
\newblock In: CVPR. 2021, 13653--13662
\bibAnnoteFile{MRAA}

\bibitem{make-a-video}
Singer U, Polyak A, Hayes T, et~al.
\newblock Make-a-video: Text-to-video generation without text-video data.
\newblock ICLR, 2023
\bibAnnoteFile{make-a-video}

\bibitem{DDIM}
Song J, Meng C, and Ermon S.
\newblock Denoising diffusion implicit models.
\newblock In: ICLR. 2021
\bibAnnoteFile{DDIM}

\bibitem{mocogan}
Tulyakov S, Liu M~Y, Yang X, et~al.
\newblock Moco{GAN}: Decomposing motion and content for video generation.
\newblock In: CVPR. 2018, 1526--1535
\bibAnnoteFile{mocogan}

\bibitem{unterthiner2018towards}
Unterthiner T, Van~Steenkiste S, Kurach K, et~al.
\newblock Towards accurate generative models of video: A new metric \& challenges.
\newblock arXiv preprint arXiv:181201717, 2018
\bibAnnoteFile{unterthiner2018towards}

\bibitem{modelscopet2v}
Wang J, Yuan H, Chen D, et~al.
\newblock Modelscope text-to-video technical report.
\newblock arXiv preprint arXiv:230806571, 2023
\bibAnnoteFile{modelscopet2v}

\bibitem{disco}
Wang T, Li L, Lin K, et~al.
\newblock Disco: Disentangled control for referring human dance generation in real world.
\newblock In: ICLR. 2024
\bibAnnoteFile{disco}

\bibitem{videocomposer}
Wang X, Yuan H, Zhang S, et~al.
\newblock Videocomposer: Compositional video synthesis with motion controllability.
\newblock NeurIPS, 2023
\bibAnnoteFile{videocomposer}

\bibitem{wang2021oadtr}
Wang X, Zhang S, Qing Z, et~al.
\newblock Oadtr: Online action detection with transformers.
\newblock In: ICCV. 2021, 7565--7575
\bibAnnoteFile{wang2021oadtr}

\bibitem{SSTAP}
Wang X, Zhang S, Qing Z, et~al.
\newblock Self-supervised learning for semi-supervised temporal action proposal.
\newblock In: CVPR. 2021, 1905--1914
\bibAnnoteFile{SSTAP}

\bibitem{wang2023few}
Wang X, Zhang S, Yuan H, et~al.
\newblock Few-shot action recognition with captioning foundation models.
\newblock arXiv preprint arXiv:231010125, 2023
\bibAnnoteFile{wang2023few}

\bibitem{wang2023videolcm}
Wang X, Zhang S, Zhang H, et~al.
\newblock Videolcm: Video latent consistency model.
\newblock arXiv preprint arXiv:231209109, 2023
\bibAnnoteFile{wang2023videolcm}

\bibitem{tft2v}
Wang X, Zhang S, Yuan H, et~al.
\newblock A recipe for scaling up text-to-video generation with text-free videos.
\newblock In: CVPR. 2024
\bibAnnoteFile{tft2v}

\bibitem{wang2020g3an}
Wang Y, Bilinski P, Bremond F, et~al.
\newblock G3an: Disentangling appearance and motion for video generation.
\newblock In: CVPR. 2020, 5264--5273
\bibAnnoteFile{wang2020g3an}

\bibitem{wang2004image}
Wang Z, Bovik A~C, Sheikh H~R, et~al.
\newblock Image quality assessment: from error visibility to structural similarity.
\newblock IEEE Transactions on Image Processing, 2004.
\newblock 13:600--612
\bibAnnoteFile{wang2004image}

\bibitem{wei2023dreamvideo}
Wei Y, Zhang S, Qing Z, et~al.
\newblock Dreamvideo: Composing your dream videos with customized subject and motion.
\newblock arXiv preprint arXiv:231204433, 2023
\bibAnnoteFile{wei2023dreamvideo}

\bibitem{tune-a-video}
Wu J~Z, Ge Y, Wang X, et~al.
\newblock Tune-a-video: One-shot tuning of image diffusion models for text-to-video generation.
\newblock In: ICCV. 2023, 7623--7633
\bibAnnoteFile{tune-a-video}

\bibitem{xing2023make}
Xing J, Xia M, Liu Y, et~al.
\newblock Make-your-video: Customized video generation using textual and structural guidance.
\newblock arXiv preprint arXiv:230600943, 2023
\bibAnnoteFile{xing2023make}

\bibitem{xing2023simda}
Xing Z, Dai Q, Hu H, et~al.
\newblock Simda: Simple diffusion adapter for efficient video generation.
\newblock arXiv preprint arXiv:230809710, 2023
\bibAnnoteFile{xing2023simda}

\bibitem{xu2024you}
Xu Z, Wei K, Yang X, et~al.
\newblock Do you guys want to dance: Zero-shot compositional human dance generation with multiple persons.
\newblock arXiv preprint arXiv:240113363, 2024
\bibAnnoteFile{xu2024you}

\bibitem{magicanimate}
Xu Z, Zhang J, Liew J~H, et~al.
\newblock Magicanimate: Temporally consistent human image animation using diffusion model.
\newblock arXiv preprint arXiv:231116498, 2023
\bibAnnoteFile{magicanimate}

\bibitem{yang2024plainmamba}
Yang C, Chen Z, Espinosa M, et~al.
\newblock Plainmamba: Improving non-hierarchical mamba in visual recognition.
\newblock arXiv preprint arXiv:240317695, 2024
\bibAnnoteFile{yang2024plainmamba}

\bibitem{yang2018pose}
Yang C, Wang Z, Zhu X, et~al.
\newblock Pose guided human video generation.
\newblock In: ECCV. 2018, 201--216
\bibAnnoteFile{yang2018pose}

\bibitem{DWpose}
Yang Z, Zeng A, Yuan C, et~al.
\newblock Effective whole-body pose estimation with two-stages distillation.
\newblock In: ICCV. 2023, 4210--4220
\bibAnnoteFile{DWpose}

\bibitem{yin2023dragnuwa}
Yin S, Wu C, Liang J, et~al.
\newblock Dragnuwa: Fine-grained control in video generation by integrating text, image, and trajectory.
\newblock arXiv preprint arXiv:230808089, 2023
\bibAnnoteFile{yin2023dragnuwa}

\bibitem{yu2023bidirectionally}
Yu W~Y, Po L~M, Cheung R~C, et~al.
\newblock Bidirectionally deformable motion modulation for video-based human pose transfer.
\newblock In: ICCV. 2023, 7502--7512
\bibAnnoteFile{yu2023bidirectionally}

\bibitem{yuan2023instructvideo}
Yuan H, Zhang S, Wang X, et~al.
\newblock Instructvideo: Instructing video diffusion models with human feedback.
\newblock arXiv preprint arXiv:231212490, 2023
\bibAnnoteFile{yuan2023instructvideo}

\bibitem{zablotskaia2019dwnet}
Zablotskaia P, Siarohin A, Zhao B, et~al.
\newblock Dwnet: Dense warp-based network for pose-guided human video generation.
\newblock arXiv preprint arXiv:191009139, 2019
\bibAnnoteFile{zablotskaia2019dwnet}

\bibitem{UBCfashion}
Zablotskaia P, Siarohin A, Zhao B, et~al.
\newblock Dwnet: Dense warp-based network for pose-guided human video generation.
\newblock arXiv preprint arXiv:191009139, 2019
\bibAnnoteFile{UBCfashion}

\bibitem{controlnet}
Zhang L, Rao A, and Agrawala M.
\newblock Adding conditional control to text-to-image diffusion models.
\newblock In: ICCV. 2023, 3836--3847
\bibAnnoteFile{controlnet}

\bibitem{zhang2022exploring}
Zhang P, Yang L, Lai J~H, et~al.
\newblock Exploring dual-task correlation for pose guided person image generation.
\newblock In: CVPR. 2022, 7713--7722
\bibAnnoteFile{zhang2022exploring}

\bibitem{DPTN}
Zhang P, Yang L, Lai J~H, et~al.
\newblock Exploring dual-task correlation for pose guided person image generation.
\newblock In: CVPR. 2022, 7713--7722
\bibAnnoteFile{DPTN}

\bibitem{zhang2018unreasonable}
Zhang R, Isola P, Efros A~A, et~al.
\newblock The unreasonable effectiveness of deep features as a perceptual metric.
\newblock In: CVPR. 2018, 586--595
\bibAnnoteFile{zhang2018unreasonable}

\bibitem{zhang2023i2vgen}
Zhang S, Wang J, Zhang Y, et~al.
\newblock I2vgen-xl: High-quality image-to-video synthesis via cascaded diffusion models.
\newblock arXiv preprint arXiv:231104145, 2023
\bibAnnoteFile{zhang2023i2vgen}

\bibitem{zhang2023controlvideo}
Zhang Y, Wei Y, Jiang D, et~al.
\newblock Controlvideo: Training-free controllable text-to-video generation.
\newblock arXiv preprint arXiv:230513077, 2023
\bibAnnoteFile{zhang2023controlvideo}

\bibitem{TPS}
Zhao J and Zhang H.
\newblock Thin-plate spline motion model for image animation.
\newblock In: CVPR. 2022, 3657--3666
\bibAnnoteFile{TPS}

\bibitem{zhao2023controlvideo}
Zhao M, Wang R, Bao F, et~al.
\newblock Controlvideo: Adding conditional control for one shot text-to-video editing.
\newblock arXiv preprint arXiv:230517098, 2023
\bibAnnoteFile{zhao2023controlvideo}

\bibitem{zhou2022magicvideo}
Zhou D, Wang W, Yan H, et~al.
\newblock Magicvideo: Efficient video generation with latent diffusion models.
\newblock arXiv preprint arXiv:221111018, 2022
\bibAnnoteFile{zhou2022magicvideo}

\bibitem{zhu2024poseanimate}
Zhu B, Wang F, Lu T, et~al.
\newblock Poseanimate: Zero-shot high fidelity pose controllable character animation.
\newblock arXiv preprint arXiv:240413680, 2024
\bibAnnoteFile{zhu2024poseanimate}

\bibitem{Visionmamba}
Zhu L, Liao B, Zhang Q, et~al.
\newblock Vision mamba: Efficient visual representation learning with bidirectional state space model.
\newblock arXiv preprint arXiv:240109417, 2024
\bibAnnoteFile{Visionmamba}

\bibitem{champ}
Zhu S, Chen J~L, Dai Z, et~al.
\newblock Champ: Controllable and consistent human image animation with 3d parametric guidance.
\newblock arXiv preprint arXiv:240314781, 2024
\bibAnnoteFile{champ}

\end{thebibliography}
\bibliographystyle{scis}
}


\end{document}